\documentclass{article} %
\usepackage[preprint]{colm2026_conference}

\usepackage{microtype}
\usepackage{hyperref}
\usepackage{url}
\usepackage{booktabs}
\usepackage{multirow}
\usepackage{caption}
\usepackage{enumitem}
\usepackage{empheq}
\usepackage{tcolorbox}
\usepackage{mdframed}
\usepackage{graphicx}
\usepackage{tcolorbox}
\usepackage{listings}

\usepackage{amsmath, amssymb, amsthm}

\newcommand{\B}{\mathrm{B}}
\newcommand{\Bet}{\mathrm{Beta}}

\tcbuselibrary{theorems}
\definecolor{approach1}{HTML}{1F77B4}  %
\definecolor{approach2}{HTML}{D62728}  %

\usepackage{lineno}

\definecolor{darkblue}{rgb}{0, 0, 0.5}
\hypersetup{colorlinks=true, citecolor=darkblue, linkcolor=darkblue, urlcolor=darkblue}

\title{Test-Time Scaling Makes Overtraining Compute-Optimal}

\author{Nicholas Roberts$^\mu$\thanks{Corresponding author: \href{mailto:nick11roberts@cs.wisc.edu}{\texttt{nick11roberts@cs.wisc.edu}}.} \,\,\,\,\, Sungjun Cho$^\mu$ \,\,\,\, Zhiqi Gao$^\mu$ \,\,\,\, Tzu-Heng Huang$^\mu$ \,\,\,\, Albert Wu$^\mu$ \,\,\, \\
\textbf{Gabriel Orlanski$^\mu$ \, Avi Trost$^\mu$ \, Kelly Buchanan$^\sigma$ \, Aws Albarghouthi$^\mu$ \, Frederic Sala$^\mu$} \\
$^\mu$University of Wisconsin-Madison \,\,\, $^\sigma$Stanford University \\
}

\begin{document}

\ifcolmsubmission
\linenumbers
\fi

\maketitle

\begin{abstract}
Modern LLMs scale at test-time, e.g. via repeated sampling, where inference cost grows with model size and the number of samples. 
This creates a trade-off that pretraining scaling laws, such as Chinchilla, do not address. 
We present Train-to-Test ($T^2$) scaling laws that jointly optimize model size, training tokens, and number of inference samples under fixed end-to-end budgets. 
$T^2$ modernizes pretraining scaling laws with pass@$k$ modeling used for test-time scaling, then jointly optimizes pretraining and test-time decisions. 
Forecasts from $T^2$ are robust over distinct modeling approaches: measuring joint scaling effect on the task loss and modeling impact on task accuracy. 
Across eight downstream tasks, we find that when accounting for inference cost, optimal pretraining decisions shift radically into the overtraining regime, well-outside of the range of standard pretraining scaling suites. 
We validate our results by pretraining heavily overtrained models in the optimal region that $T^2$ scaling forecasts, confirming their substantially stronger performance compared to pretraining scaling alone. 
Finally, as frontier LLMs are post-trained, we show that our findings survive the post-training stage, making $T^2$ scaling meaningful in modern deployments. 
\end{abstract}

\section{Introduction}

Pretraining scaling laws tell us how to optimally train language models, but not how to deploy them~\citep{kaplan2020scaling, hoffmann2022training}. 
Test-time scaling laws tell us how to optimally allocate compute at deployment, but not how to train models~\citep{snell2024scaling, brown2025large}.
The two have developed largely in \textit{isolation}, yet are fundamentally coupled. Model size and training duration determine both the quality and cost of inference samples. 
Models designed to reason through frontier research problems will be sampled from hundreds or thousands of times~\citep{jaech2024openai, guo2025deepseek}; these should be trained differently from chat models that instantly answer everyday questions. 

\textbf{Should parameter and token counts change if you know how your model will be used at test time?}
In practice, Chinchilla~\citep{hoffmann2022training} scaling laws guide the allocation of pretraining compute for flagship models.
However, modern model releases are families spanning a range of sizes~\citep{touvron2023llama, groeneveld2024olmo, qwen2024qwen2}, with the lower end intentionally \textit{overtrained} well beyond Chinchilla-optimal ratios to reduce per-query inference cost.
This makes them natural candidates for test-time scaling, yet nothing connects pretraining decisions to this inference strategy.
\textbf{No existing scaling law captures the core tradeoff}: smaller models are cheaper per sample but weaker per sample, and the benefit of repeated sampling is a highly nonlinear function of per-sample quality.

Unifying pretraining and inference scaling is challenging because the two regimes operate under fundamentally different evaluation criteria.
Pretraining is evaluated using the loss, a smooth, continuous quantity. 
Test-time scaling, by contrast, is evaluated through downstream task metrics such as pass@$k$---the probability of producing at least one correct answer in $k$ independent attempts.
Should a unified scaling law across pretraining and test-time scaling model the loss or model the pass@$k$ accuracy? 

Prior work has addressed pieces of this problem but not the whole. 
\citet{sardana2024beyond} extends Chinchilla to account for inference cost, but considers only the aggregate volume of \textit{single-pass} serving instead of the multiplicative cost and performance gains from repeated sampling.
Recent studies empirically show that allocating more inference compute to smaller models via repeated sampling can match or exceed the performance of larger ones~\citep{brown2025large, snell2024scaling}, but they treat pretrained models as given and do not address how they should have been trained. 
\citet{schaeffer2025pretraining} develop scaling laws that predict pass@$k$ from pretraining compute, but treat this as forecasting rather than an optimization problem---they predict what performance \emph{will be} for a given model, not what model \emph{should be} trained for a given budget. 
No existing work jointly optimizes model size, training duration, and the number of inference samples under a single compute budget.

In this work, we close the loop between pre-training and test-time scaling. 
We propose \textbf{Train-to-Test ($T^2$) scaling laws} that predict performance as a function of model size $N$, training tokens $D$, and number of samples $k$, and optimize over all three under a total compute budget that includes both training ($6ND$) and inference ($2Nk$) cost. 
Following Chinchilla, we evaluate multiple modeling approaches: whether to model the loss or pass@$k$ as functions of $N$, $D$, and $k$.
Although the two approaches are quite different, we find that they agree closely: both suggest substantial overtraining and test-time scaling across our evaluations.
We build on an existing set of Chinchilla scaling checkpoints from \citet{porian2024resolving}, extending it into the overtrained regime and assembling a testbed of over 100 models across 12 compute levels spanning three orders of magnitude.

\begin{figure}[t!]
    \vspace{-30pt}
    \begin{center}
    \includegraphics[width=0.8\linewidth]{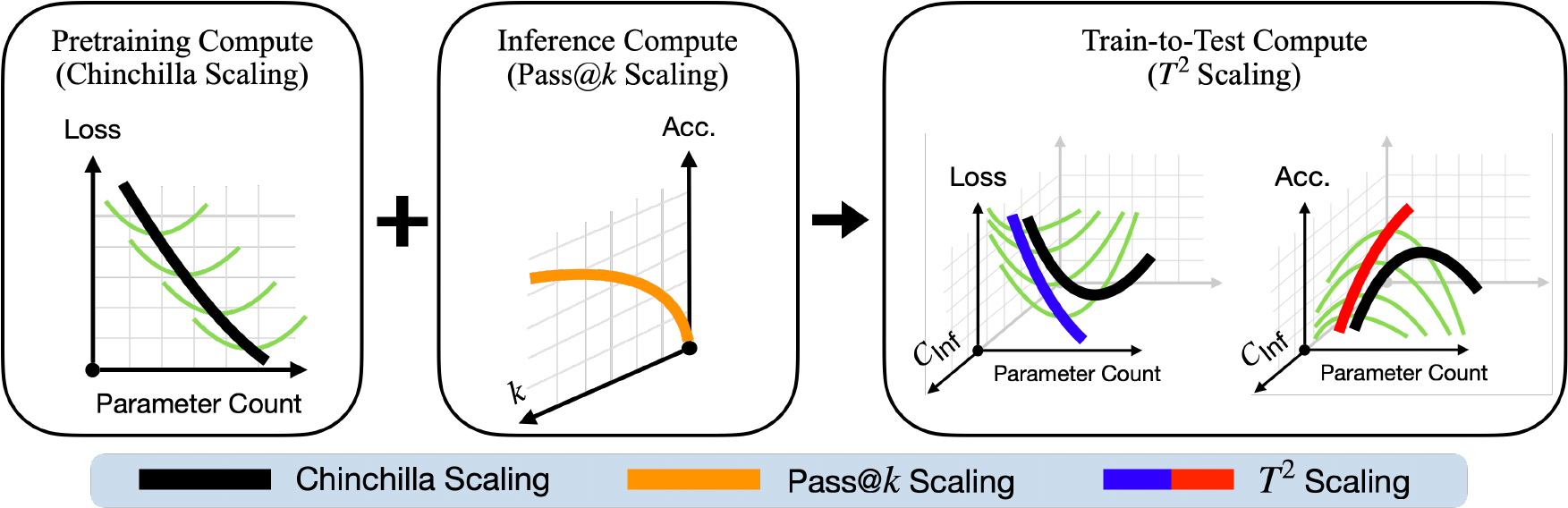}
    \end{center}
    \centering
    \vspace{-8pt}
    \caption{Our $T^2$ scaling laws combine Chinchilla scaling for pretraining with pass@$k$ modeling for test-time scaling via repeated sampling to obtain optimal pretraining allocations subject to a test-time scaling budget. $T^2$ recommends overtraining compared to Chinchilla. 
    }
    \label{fig:framework}
    \vspace{-1em}
\end{figure}

Using $T^2$ scaling laws, we find that \emph{\textbf{optimal pretraining decisions shift radically into the overtraining regime}} when considering test-time compute. 
When we correct for the cost of repeated sampling, the optimal model is substantially smaller and more overtrained than what Chinchilla prescribes. 
Our evaluation spans eight tasks covering knowledge, reasoning, and language understanding, on which we investigate three research questions: 
\begin{enumerate}[label=\textbf{RQ\arabic*}, leftmargin=*, itemsep=1pt, topsep=3pt, parsep=0pt]
    \item \textbf{Should pretraining change if you know your test-time scaling budget?} Yes---$T^2$ scaling consistently recommends small overtrained models. (\S\ref{sec:rq1})
    \item \textbf{Does $T^2$ extrapolate to overtrained checkpoints?} Yes---we overtrain models from scratch and show that they consistently outperform Chinchilla checkpoints. (\S\ref{sec:rq2})
    \item \textbf{Does $T^2$ scaling survive post-training?} Yes---we find that compute-optimal trade-offs derived from base models persist after supervised fine-tuning. (\S\ref{sec:rq3})
\end{enumerate}

To answer these questions, we make the following contributions:
\begin{tcolorbox}[
  title=Contributions,
  colback=gray!5,
  colframe=black!60,
  boxrule=0.4pt,
  arc=2pt,
  left=6pt,
  right=6pt,
  top=0pt,
  bottom=0pt
]
\begin{itemize}[leftmargin=*, itemsep=1pt, topsep=3pt, parsep=0pt]
    \item \textbf{End-to-end scaling:} We formalize train-to-test scaling as a joint optimization over model size $N$, dataset size $D$, and inference compute $k$ under train and test budgets.
    \item \textbf{Loss and accuracy scaling:} We introduce two complementary approaches: (i) loss- and (ii) accuracy-based formulations that explicitly incorporate inference cost.    
    \item \textbf{Validation on overtrained checkpoints:} We train models in the predicted overtrained regime and show improved performance under a range of fixed inference budgets.
    \item \textbf{Interactions with post-training:} The predictions from our scaling approach persist after post-training, even though overtrained models are harder to fine-tune. 
\end{itemize}
\end{tcolorbox}

\section{Background}
Our work connects two important areas: (i) pretraining scaling laws and (ii) test-time sampling strategies after deployment. 
We begin with their setups then dive into our new modeling techniques. 
A summary of additional related work can be found in Appendix~\ref{app:related_works}. 

\textbf{Chinchilla scaling laws for pretraining.}
The Chinchilla scaling law~\citep{hoffmann2022training} models the pretraining loss as a function of finite model capacity $N$ and dataset size $D$ (number of training tokens): $L(N, D) = E + \frac{A}{N^{\alpha}} + \frac{B}{D^{\beta}}$, 
where $E$ represents an irreducible loss floor fit for the given data distribution and evaluation setup while the remaining terms capture reducible contributions from $N$ and $D$. 
The parameters $A$, $B$, $\alpha$, $\beta$, and $E$ are all non-negative and are fit empirically from a grid of training runs. 
Here, the loss is assumed to be the negative log-likelihood (NLL) over the data distribution: $\mathbb{E}_{(x, y)\sim\mathcal{D}}[-\log(p(y|x))]$ with $p(y|x)$ being the probability assigned by the model. 
Given a pretraining budget $C_{\text{train}} \approx 6ND$, the \emph{compute-optima} minimize $L$ subject to this constraint, 
yielding $N^{*}(C_{\text{train}}) \propto C_{\text{train}}^{a}$ and $D^{*}(C_{\text{train}}) \propto C_{\text{train}}^{b}$ with $a \approx b \approx 0.5$. 
That is, the optimal model size and training tokens should scale at similar rates as a function of the pretraining compute budget.

\textbf{Pass@k estimation for test-time scaling.}
The standard metric for evaluating repeated sampling is pass@$k$: draw $k$ independent samples from a model and succeed if \emph{any} sample is correct.
For a single problem $i$ with per-sample success probability $p_i$, the probability of at least one answer in $k$ attempts being correct is $\text{pass@}k_i = 1 - (1 - p_i)^k$. 
Aggregating over a benchmark $\mathcal{D}$ of $M$ problems gives the expected pass@$k$:
\begin{equation*}
    \text{pass@}k_{\mathcal{D}} = \mathbb{E}_{i \sim \mathcal{D}}\left[ \text{pass@}k_i\right] = \dfrac{1}{M} \sum_{i=1}^M \left[1 - (1 - p_i)^k \right]. \label{eqn:expected_pass_at_k}
\end{equation*}

\section{Estimating Optimal Pretraining Allocations for Test-Time Scaling}
We present two modeling approaches for $T^2$ scaling that answer our central research question: \textbf{{should choices made during pretraining change if you know your test-time scaling budget?}} 
In our first approach, we model the impact of repeated sampling on the loss by fitting a parametric function of the negative log pass@$k$. 
In our second approach, we model the pass@$k$ accuracy directly by composing Chinchilla scaling with a pass@$k$ estimator. 
In \S\ref{sec:res}, we show that our findings are robust across both approaches.  
Finally, once we establish these two approaches, we answer our main research question by standardizing the test-time scaling budget: using more repeated samples for smaller models and fewer for larger models. 
Standardizing the inference budget of test-time scaling across checkpoints allows us to see how optimal pretraining decisions shift in light of test-time scaling considerations.
\textbf{If the optimal pretraining decisions (model size and the number of training tokens) shift compared to those recommended by standard Chinchilla scaling, then the answer to RQ1 is yes: pretraining decisions should change if you know your test-time scaling budget.}

We first describe the optimization objectives of our $T^2$ approaches. 
Given a compute budget for training ($C_{\text{train}}$) and inference ($C_{\text{inf}}$), the optimization problem in terms of the NLL is:
\begin{equation}
    \min_{N,D, k} \;\; L(N, D, k) \qquad \text{s.t.} \quad 6ND \leq C_{\text{train}} \,\, \text{ and } \,\, 2Nk \leq C_{\text{inf}},
    \label{eq:joint_opt_1}
\end{equation}
or similarly, in terms of the pass@$k$ accuracy: 
\begin{equation}
    \max_{N,D, k} \;\; \text{Acc}(N, D, k) \qquad \text{s.t.} \quad 6ND \leq C_{\text{train}} \,\,\text{ and } \,\,  2Nk \leq C_{\text{inf}}.
    \label{eq:joint_opt_2}
\end{equation}
$L(N,D,k)$ and $\text{Acc}(N, D, k)$ represent the aggregated NLL and accuracy respectively, as functions of model capacity $N$, dataset size $D$, and number of sampling attempts $k$.

\subsection{\textcolor{blue}{Approach 1}: $T^2$ as a Parametric Model of the Task Loss}
Our first approach models the loss as a function of the parameter count $N$, training tokens $D$, and the number of repeated samples $k$ used at test-time in order to optimize Equation~\ref{eq:joint_opt_1}. 
First, in order to make repeated sampling compatible with the negative log likelihood (NLL), we rewrite the single-sample probability in terms of the probability that the target outcome is obtained at least once under $k$ repeated samples, following prior work on pass@$k$~\citep{chen2021evaluating,brown2025large,ehrlich2025codemonkeys,schaeffer2025how}. 
That is, working with the definition of $\text{pass@}k_i$ allows us to define the corresponding NLL-style objective under repeated sampling as
\begin{equation*}
    \mathbb{E}_{i \sim \mathcal{D}_{\text{task}}}[-\log \text{pass@}k_i] = \mathbb{E}_{i \sim \mathcal{D}_{\text{task}}}\left[-\log \left(1 - (1 - p_i)^k\right)\right],
\end{equation*}

where $\mathcal{D}_{\text{task}}$ is a distribution over samples $i$ representing a downstream task.

With this in place, we can model the negative log pass@$k$ as an extension of the Chinchilla scaling law, $\widehat{L}(N, D)$ by adding a power-law term in $k$: 
\begin{equation*}
    \widehat{L}(N, D, k) = \widehat{L}(N, D) + \frac{G}{k^{\gamma}} = E + \frac{A}{N^{\alpha}} + \frac{B}{D^{\beta}} + \frac{G}{k^{\gamma}}. 
\end{equation*}

We choose this model because prior work has found that the negative log pass@$k$ contribution from $k$ yields power law scaling\footnote{By Jensen's inequality, our NLL-style objective acts as an upper-bounding surrogate on the negative log expected pass@$k$, which scales as a power law (we minimize the expected negative log pass@$k$). Therefore, minimizing our surrogate minimizes the quantity of interest.} under an assumption that the task difficulty distribution can be modeled by a Beta distribution, which has been found to hold in practice~\citep{brown2025large, schaeffer2025how}.
This has convenient properties when combined with the other power law terms in $N$ and $D$ in the Chinchilla scaling law: 

First, when $k=1$, we recover standard Chinchilla scaling: 
\begin{equation*}
\widehat{L}(N, D, 1) = E' + \frac{A}{N^{\alpha}} + \frac{B}{D^{\beta}} = \widehat{L}(N, D),
\end{equation*}
where $E' = E + G$ absorbs the additional constant. 
Second, a property of Chinchilla scaling is that as $N, D \to \infty$, the model approaches the `irreducible loss' term $E$. 
Given its power law form, this is still true when $k$ approaches infinity alongside $N$ and $D$. 

\subsection{\textcolor{red}{Approach 2}: $T^2$ as a Parametric Model of the Task Accuracy}
While the previous model is simple, it trades off interpretability---practitioners often value pass@$k$ forecasts due to their interpretation as the likelihood of solving a problem given a certain compute investment. 
Our second approach addresses this by modeling the pass@$k$ directly as an accuracy-like metric as a function of $N$, $D$, and $k$, which optimizes Equation~\ref{eq:joint_opt_2}. 

A naive approach to modeling pass@$k$ might be to begin with $\widehat{L}(N, D)$, and simply map the NLL to accuracy $p$ for the same task, then compute $\text{pass@}k = 1 - (1 - p)^k$. 
Prior work has shown that the relationship between the mean NLL and the mean accuracy can be well approximated using a fitted sigmoid~\citep{llama3}. 
In other words, we can model the mean single-pass task accuracy, $\mathbb{E}_{\mathcal{D}_{\text{task}}}[\text{Acc}(N, D)]$, as $\sigma_{\theta}(\widehat{L}(N, D))$ with a parameterized sigmoid $\sigma_{\theta}$ fit to pairs of NLL and accuracy values on the task distribution across the model population. 
So this naive model of the pass@$k$ might take the following form:
\begin{equation*}
    \widehat{\text{Acc}}_{\text{naive}}(N, D, k) = 1 - (1 - \sigma_{\theta}(L(N, D)))^k. 
\end{equation*}
However, our goal is instead to obtain an estimator of the \textit{mean pass@$k$ accuracy}, $\mathbb{E}_{\mathcal{D}_{\text{task}}}[\text{Acc}(N, D, k)]$ that depends on the scaling parameters, rather than the single-pass accuracy, so this naive model overestimates due to the concavity of the pass@$k$: 
\begin{align*}
    1 - (1 - \mathbb{E}_{\mathcal{D}_{\text{task}}}[\text{Acc}(N, D)])^k &\geq \mathbb{E}_{\mathcal{D}_{\text{task}}}[1 - (1-\text{Acc}(N, D))^k] \\ 
    &= \mathbb{E}_{\mathcal{D}_{\text{task}}}[\text{Acc}(N, D, k)]. 
\end{align*}

A simple way to avoid overestimating the pass@$k$ would be to directly use the per-question probabilities from model likelihoods, which would allow us to compute the mean pass@$k$ exactly. 
However, our goal is a scaling law, a parametric model that can forecast pass@$k$ at unevaluated $(N, D, k)$ configurations. 
This requires us to model the \textit{distribution} of per-question probabilities and how this distribution varies with model size and training tokens. 

Intuitively, we want to account for the natural spread of difficulty between tasks in our data distribution.
We do this by modeling the per-question single-pass accuracies as a Beta distribution, following prior work~\citep{kazdan2025efficient}.
We model $\text{Acc}(N, D) \sim \Bet(a_{N, D}, b_{N, D})$,
and parameters $a_{N,D}$ with $b_{N,D}$ related to $N$ and $D$ via the NLL, which we model as a Beta regression problem. 
Using the mean ($\mu$) and sample size ($\nu$) parameterization of the Beta distribution, we model $\mu \in (0, 1)$ and $\nu \in (0, \infty)$ using standard link functions from Beta regression: a logit link for the mean (which we rescale with an additional parameter), and a log link for the sample size. 
We relate this to the loss by using the Chinchilla loss estimate as our linear predictor. 
This yields the following parameterization of $a_{N,D}$ and $b_{N,D}$: 
\begin{align*}
    \mu_{N, D} &= \sigma_{\theta}(\widehat{L}(N, D)) = \frac{\theta_2}{1 + \exp\bigl(\theta_1 \cdot (\widehat{L}(N, D) - \theta_0)\bigr)}, \\
    \nu_{N, D} &= \exp(\theta_3 + \theta_4 \cdot \widehat{L}(N, D)), \\
    a_{N, D} &= \mu_{N, D} \nu_{N, D}, \\
    b_{N, D} &= (1 - \mu_{N, D}) \nu_{N, D}.
\end{align*}

Finally, using this model of the single-pass accuracy, we obtain the following pass@$k$ model via properties of the Beta distribution:\footnote{$\B(a,b)=\frac{\Gamma(a)\Gamma(b)}{\Gamma(a+b)}$ is the Beta function, where $\Gamma$ is the Gamma function.} 
\begin{align*}
     \widehat{\text{Acc}}(N, D, k)  &= \mathbb{E}_{\text{Acc}(N, D)  \sim \Bet(a_{N,D},b_{N,D})}\bigl[1 - (1-\text{Acc}(N, D) )^k\bigr] \\
             &= 1 - \mathbb{E}_{\text{Acc}(N, D) \sim \Bet(a_{N,D},b_{N,D})}\bigl[(1-\text{Acc}(N, D) )^k\bigr] \\
             &= 1 - \frac{\B(a_{N,D},\, b_{N,D} + k)}{\B(a_{N,D},\, b_{N,D})} \\
             &= 1 - \frac{\B(\mu_{N, D} \nu_{N, D},\, (1 - \mu_{N, D}) \nu_{N, D} + k)}{\B(\mu_{N, D} \nu_{N, D},\, (1 - \mu_{N, D}) \nu_{N, D})}.
\end{align*}

\subsection{Inference Cost Correction}
We equalize our $T^2$ scaling laws over an inference budget, $C_{\text{inf}} $, measured as the inference FLOPs per-token served. 
Just as the pretraining cost, $C_{\text{train}} = 6 N D$, scales multiplicatively as a function of $N$ and the number of training tokens $D$, the inference budget $C_{\text{inf}} $ scales multiplicatively in $k$ and approximately $2N$ FLOPs for a forward pass: 
\begin{equation*}
    C_{\text{inf}} = 2 N k. 
\end{equation*}
Then for a fixed budget $C_{\text{inf}}$, this gives us  
\begin{equation*}
    k = \frac{C_{\text{inf}}}{2 N}, 
\end{equation*}
where smaller models are allocated more repeated samples compared to larger models, subject to the same inference budget. 
We plug this into both of our $T^2$ scaling approaches, which gives us our inference-corrected loss model:\footnote{Optimization details for fitting \textcolor{blue}{Approach~1} and \textcolor{red}{Approach~2} can be found in Appendix~\ref{app:t^2_scaling}.}
\begin{tcolorbox}[
  title=Approach 1,
  colback=white,
  colframe=blue,
  fonttitle=\bfseries\small,
  boxrule=0.5pt
]
    \[
    \widehat{L}\left(N, D, \frac{C_{\text{inf}}}{2 N}\right) = \widehat{L}(N, D) + \frac{G}{k^{\gamma}} = E + \frac{A}{N^{\alpha}} + \frac{B}{D^{\beta}} + \frac{G}{\left(\frac{C_{\text{inf}}}{2 N}\right)^{\gamma}}, 
    \]
\end{tcolorbox}
and our inference-corrected pass@$k$ accuracy model:
\begin{tcolorbox}[
  title=Approach 2,
  colback=white,
  colframe=red,
  fonttitle=\bfseries\small,
  boxrule=0.5pt
]
    \[
    \widehat{\text{Acc}}\left(N, D, \frac{C_{\text{inf}}}{2 N}\right) = 1 - \frac{\B(\mu_{N, D} \nu_{N, D},\, (1 - \mu_{N, D}) \nu_{N, D} + \frac{C_{\text{inf}}}{2 N})}{\B(\mu_{N, D} \nu_{N, D},\, (1 - \mu_{N, D}) \nu_{N, D})}.
    \]
\end{tcolorbox}

Now for both models, we can choose an inference budget $C_{\text{inf}}$, and observe the pretraining decisions that optimize both the pretraining and inference budgets $C_{\text{train}}$ and $C_{\text{inf}}$. 
We represent \textcolor{blue}{Approach 1} in blue and \textcolor{red}{Approach 2} in red for consistency with our Figures.

\section{Experiments}\label{sec:res}
In this section, we provide experimental results addressing the three research questions about our $T^2$ scaling approaches.%
First, in \S\ref{sec:rq1}, we show that if you know your test-time scaling budget prior to pretraining, you should overtrain significantly beyond the standard Chinchilla recommendation of 20 tokens per parameter. 
In \S\ref{sec:rq2}, we validate our predictions against overtrained checkpoints that extend standard Chinchilla scaling suites, showing that our scaling approaches extrapolate to the optimal regions that they predict. 
Finally, in \S\ref{sec:rq3}, we show that overtraining predictions from our $T^2$ approaches persist after post-training. 
We fit $T^2$ scaling to checkpoints from~\citet{porian2024resolving}, which we extend with additional overtrained checkpoints, all trained on RefinedWeb~\citep{refinedweb}. 

\textbf{Tasks.}
We evaluate $T^2$ across eight real and synthetic tasks that we select to be simple enough for small base models, as all of our checkpoints have fewer than 1B parameters. 
The real tasks that we evaluate include the OpenAI variant of LAMBADA~\citep{lambada,radford2019language}, ARC-Easy~\citep{ai2arc}, SciQ~\citep{SciQ}, and OpenBookQA~\citep{OpenBookQA2018}. 
We also evaluate on four synthetic tasks: simple knowledge recall, multi-step arithmetic reasoning, commonsense causal reasoning, and spatial reasoning, each consisting of 1,000 fill-in-the-blank or short completion questions that were generated using GPT-5 and Claude Opus 4.6. 
We provide additional task details in Appendix~\ref{app:evaluation_task}. 
Unless otherwise noted, we present macro averaged results over all tasks. 

\begin{figure}[t!]
    \begin{center}
    \begin{tabular}{ccc}
        \includegraphics[width=1.0\textwidth]{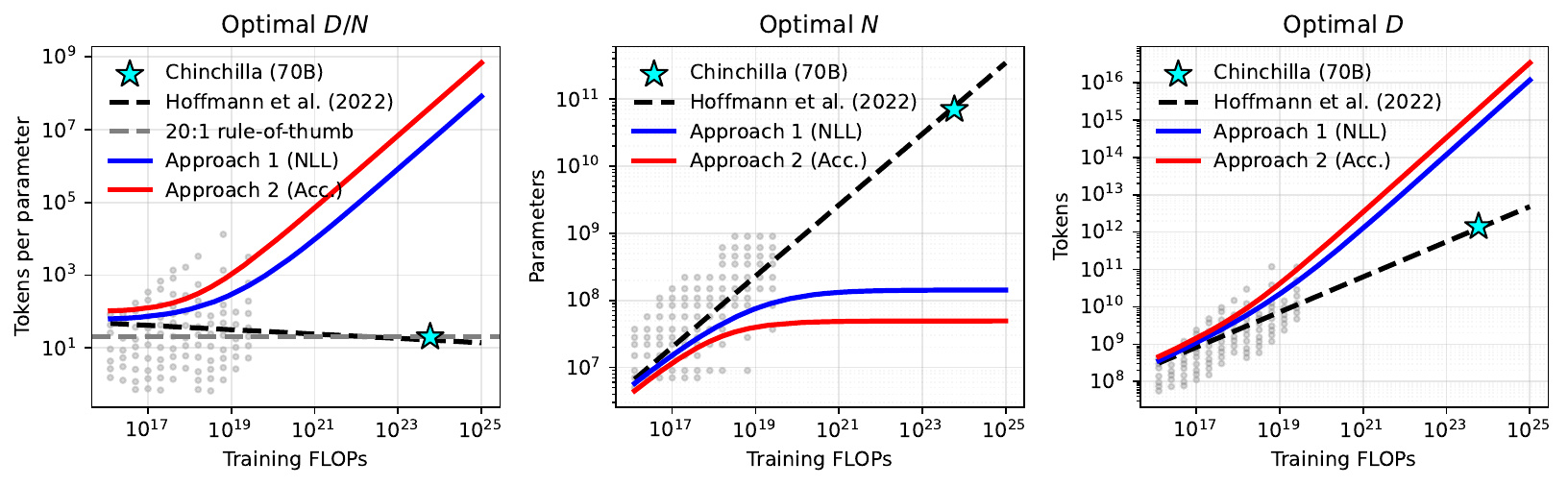}
    \end{tabular}
    \end{center}
    \centering
    \vspace{-8pt}
    \caption{
    Optimal pretraining forecasts predicted by both $T^2$ approaches, compared to \citet{hoffmann2022training}. \textbf{(Left)} Optimal tokens per parameter (including the 20 tokens per parameter rule of thumb used by practitioners), \textbf{(Middle)} Optimal model sizes. \textbf{(Right)} Optimal training set sizes. Both $T^2$ approaches forecast extreme overtraining. 
    }
    \label{fig:tpp_comp}
\end{figure}

\subsection{RQ1: Should Pretraining Change if You Know Your Test-Time Scaling Budget?}\label{sec:rq1}
We evaluate \textbf{RQ1} by comparing the predictions from $T^2$ to Chinchilla scaling and find that if you know your test-time scaling budget, you should significantly overtrain. 

\textbf{Setup.} 
We fit both $T^2$ approaches to a suite of 106 checkpoints ranging in size from 5M to 901M parameters trained on roughly 50M to 120B tokens. 
Next, we set the per-token inference budget $C_\text{inf} = 140\text{B}$ FLOPs, or approximately the cost of a single forward pass using the 70B Chinchilla model~\citep{hoffmann2022training}. 
Finally, to compare $T^2$ forecasts to Chinchilla, we extrapolate the predictions from our $T^2$ approaches and standard Chinchilla scaling beyond our scaling suite to $10^{25}$ FLOPs. 
Using the same fits, we visualize pretraining isoFLOP profiles for both approaches. 
We compare the standard single-pass setting ($k{=}1$) to the inference-corrected setting with $C_\text{inf} = 2 \times 10^9$ FLOPs and $k = \frac{C_\text{inf}}{2N}$. 
Each of the 12 isoFLOP curves traces out a fixed pretraining budget $C_\text{train}$ by varying $N$ and $D$ subject to $C_\text{train} = 6ND$. 
We plot the Chinchilla optimal frontier in black and that of $T^2$ in red. 
Results are macro averaged across all eight tasks. 
Individual scaling fits for each task across different budgets can be found in Appendix~\ref{app:per_task_analysis}. 

\begin{figure}[t!]
    \begin{center}
    \begin{tabular}{ccc}
        \includegraphics[width=1.0\linewidth]{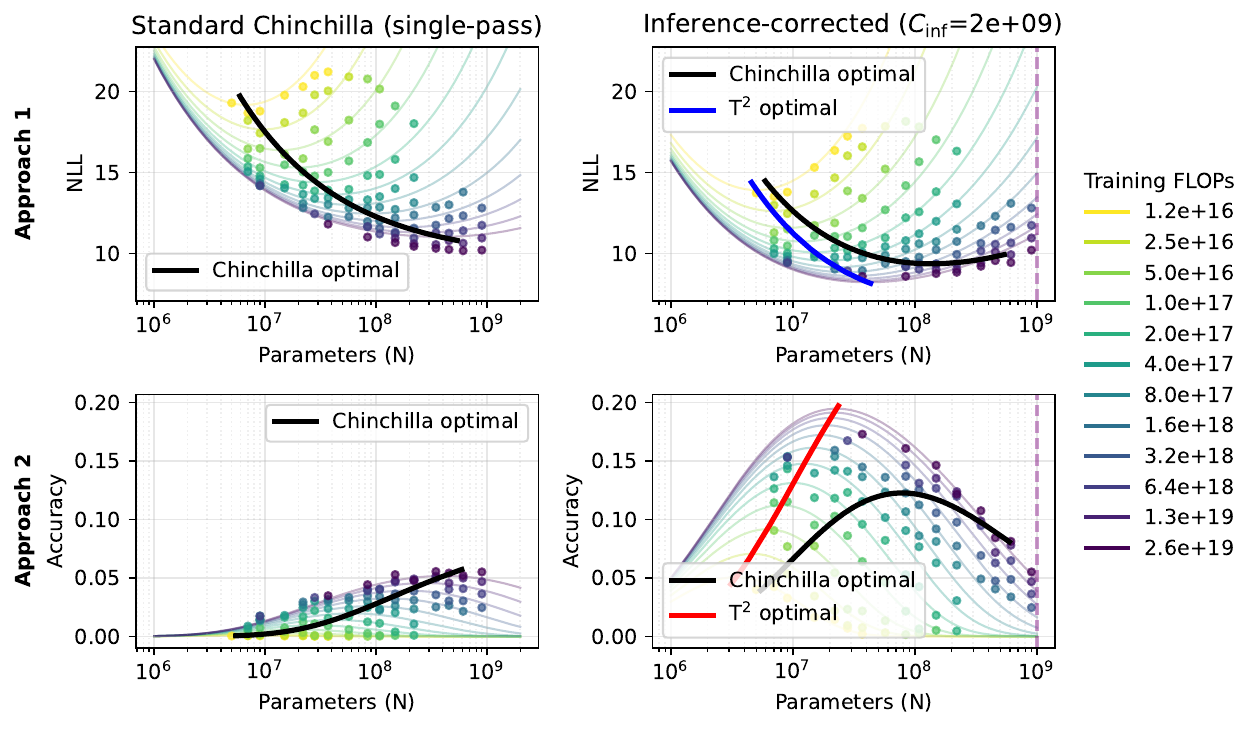}
    \end{tabular}
    \end{center}
    \centering
    \vspace{-8pt}
    \caption{$T^2$ scaling across all of our evaluation tasks. Both approaches improve monotonically over Chinchilla scaling, while Chinchilla exhibits non-monotonic scaling in $C_\text{train}$. 
    }
    \label{fig:four_panel_avg}
\end{figure}

\textbf{Results.} 
Our results are shown in Figure~\ref{fig:tpp_comp} and Figure~\ref{fig:four_panel_avg}. 
Figure~\ref{fig:tpp_comp} shows that we can answer \textbf{RQ1} in the affirmative: both $T^2$ approaches forecast models that are \textbf{dramatically smaller and more overtrained} than what Chinchilla prescribes. 
We additionally confirm that the Chinchilla scaling fit is consistent with~\citet{hoffmann2022training} by overlaying the 70B Chinchilla hero run model described in their paper, alongside the 20 tokens per parameter rule of thumb. 
Despite modeling fundamentally different quantities (NLL vs accuracy), both $T^2$ recommend extreme overtraining, with \textcolor{red}{Approach~2} recommending more aggressive overtraining than \textcolor{blue}{Approach~1}. 
Figure~\ref{fig:four_panel_avg} shows isoFLOP curves under our $T^2$ approaches, how the overtraining trend develops within our scaling population. 
At every compute scale, the optimal frontier of both $T^2$ approaches shifts considerably toward smaller overtrained models with more repeated samples compared to the Chinchilla optimum. 
When inference-corrected, we see that the Chinchilla optimal frontier exhibits non-monotonic improvement in $C_{\text{train}}$. 
This is consistent with the findings of \citet{snell2024scaling}, showing that smaller models with more test-time compute can outperform larger models. 
On the other hand, $T^2$ shows both stronger \textit{and consistently monotonic} improvement, as we jointly model pretraining and test-time scaling. 
\textbf{These results confirm that if you know your test-time scaling budget, you should substantially overtrain compared to Chinchilla optimal pretraining.}

\subsection{RQ2: Does $T^2$ Scaling Extrapolate to Overtrained Checkpoints?}\label{sec:rq2}

Next, we evaluate \textbf{RQ2} by fitting both $T^2$ approaches to standard Chinchilla scaling checkpoints and measuring the performance of extrapolation to overtrained checkpoints. 

\textbf{Setup.} 
We fit both of our $T^2$ approaches to a suite of 85 Chinchilla scaling checkpoints from \citet{porian2024resolving} (which stop short of the optimal overtraining regime that $T^2$ predicts) and measure the relative absolute error of extrapolating the predictions to 21 overtrained checkpoints that we train using an identical pretraining setup. 
We include training details and the exact checkpoint grid in Appendix~\ref{app:pretraining_details}. 
We also compare the empirical best overtrained checkpoint (among our 21) in the inference-corrected regime and compare it to the empirical Chinchilla optimal checkpoint at a pretraining budget of $C_\text{train} = 2.56 \times 10^{19}$ across all eight tasks. 
We set $C_\text{inf} = 2 \times 10^9$ for all of the above. 

\begin{figure}[t!]
\centering
\begin{minipage}[c]{0.25\linewidth}
    \centering
    \includegraphics[width=1.0\linewidth]{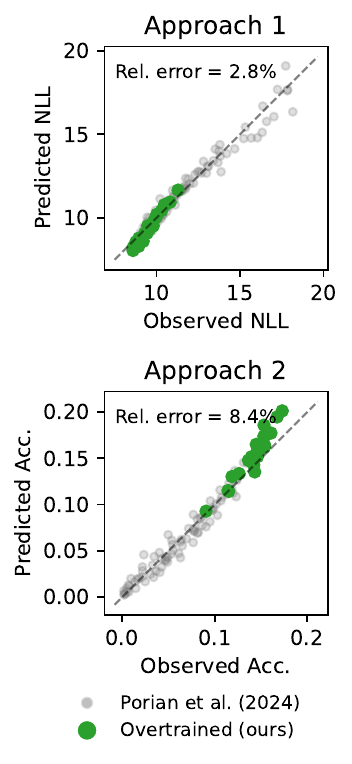}
    \captionof{figure}{Extrapolating \citet{porian2024resolving} checkpoints to the overtraining regime. }
    \label{fig:rq2_extrap}
\end{minipage}%
\hfill
\begin{minipage}[c]{0.7\linewidth}
    \centering
    \small
    \resizebox{\columnwidth}{!}{
    \begin{tabular}{r|r|ll}
    \toprule
    && Best overtrained & Chinchilla opt. \\
    \midrule
    \multirow{4}{*}{\rotatebox{90}{\textbf{Real}}} & LAMBADA OpenAI & {\bf 49.90\% (37M)} & 27.30\% (455M) \\
    & OpenBookQA & {\bf \phantom{0}1.40\% (37M)} & \phantom{0}0.30\% (901M) \\
    & SciQ & {\bf \phantom{0}1.20\% (37M)} & \phantom{0}0.22\% (611M) \\
    & ARC-Easy & {\bf \phantom{0}0.14\% (149M)} & \phantom{0}0.07\% (611M) \\
    \midrule
    \multirow{4}{*}{\rotatebox{90}{\textbf{Synthetic}}}& Simple Knowledge & {\bf 14.60\% (84M)} & \phantom{0}5.80\% (901M) \\
    & Simple Reasoning & {\bf 57.90\% (37M)} & 18.40\% (901M) \\
    & Commonsense Causal & {\bf \phantom{0}8.10\% (37M)} & \phantom{0}1.40\% (901M) \\
    & Spatial Reasoning & {\bf \phantom{0}6.00\% (37M)} & \phantom{0}1.10\% (901M) \\
    \bottomrule
    \end{tabular}
    }
    \captionof{table}{Comparison of overtrained base models vs Chinchilla optimal pass@$k$, subject to $C_{\text{train}} = 2.56 \times 10^{19}$ and $C_{\text{inf}} = 2 \times 10^{9}$ FLOPs. Optimal model sizes are shown in parentheses.}
    \label{tab:base_head_to_head}
    \vspace{2em}
    \centering
    \small
    \resizebox{\linewidth}{!}{
    \begin{tabular}{r|r|lll}
    \toprule
    & & OpenBookQA & SciQ & ARC-Easy \\
    \midrule
    \multirow{2}{*}{\rotatebox{90}{\textbf{FT}}}
    & Best overtrained & {\bf 2.80\% (37M)} & {\bf 56.10\% (149M)} & {\bf 5.60\% (149M)} \\
    & Chinchilla opt. & 0.45\% (901M) & 29.00\% (901M) & 1.50\% (901M) \\
    \midrule
    \multirow{2}{*}{\rotatebox{90}{\textbf{SFT}}}
    & Best overtrained & {\bf 2.60\% (37M)} & {\bf 66.80\% (84M)} & {\bf 8.20\% (37M)} \\
    & Chinchilla opt. & 0.38\% (901M) & 57.60\% (347M) & 3.40\% (455M) \\
    \bottomrule
    \end{tabular}
    }
    \captionof{table}{Post-training comparison of overtraining vs Chinchilla optimal pass@$k$, subject to $C_{\text{train}} = 2.56 \times 10^{19}$ and $C_{\text{inf}} = 2 \times 10^{9}$ FLOPs. Optimal model sizes are shown in parentheses.}
    \label{tab:head_to_head}
\end{minipage}
\end{figure}

\textbf{Results.} 
Our extrapolation results are shown in Figure~\ref{fig:rq2_extrap} and empirical checkpoint pass@$k$ results are shown in Table~\ref{tab:base_head_to_head}. 
Figure~\ref{fig:rq2_extrap} shows that our $T^2$ approaches both extrapolate to the 16 new overtrained checkpoints. 
While both approaches somewhat overestimate performance, \textcolor{blue}{Approach~1} extrapolates better than \textcolor{red}{Approach~2}, with a relative error of 2.8\% compared to 8.4\%. 
Table~\ref{tab:base_head_to_head} shows that our best small overtrained checkpoints always outperform the Chinchilla optimal checkpoints when inference corrected, across all eight tasks. 
\textbf{This confirms that $T^2$ extrapolates to real overtrained checkpoints, and that this phenomenon is not just an artifact of our $T^2$ approaches.}

\subsection{RQ3: Does $T^2$ Scaling Survive Post-Training?}\label{sec:rq3}
Finally, we evaluate \textbf{RQ3} by showing that our findings persist after post-training. 

\textbf{Setup.} 
We explore two canonical post-training techniques: standard fine-tuning (FT) and supervised fine-tuning (SFT), where we only fine-tune on the targets. 
We post-train on the three real tasks that have a standard training set: ARC-Easy, SciQ, and OpenBookQA, and report improved performance on the test sets for each of these. 
Additional post-training details can be found in Appendix~\ref{app:posttraining_details}. 
We allocate the same number of training steps to each checkpoint, rather than scaling training based on FLOPs, since we ultimately train to convergence. 
After post-training, we fit both $T^2$ approaches to the FT and SFT checkpoints and evaluate their optimal tokens per parameter frontier compared to base models under $T^2$ scaling and the Chinchilla frontier. 
Finally, like in \textbf{RQ2}, we compare the best overtrained FT and SFT checkpoints to the Chinchilla optimal checkpoints for each task. 

\begin{figure}[t!]
    \begin{center}
    \begin{tabular}{ccc}
        \includegraphics[width=1.0\linewidth]{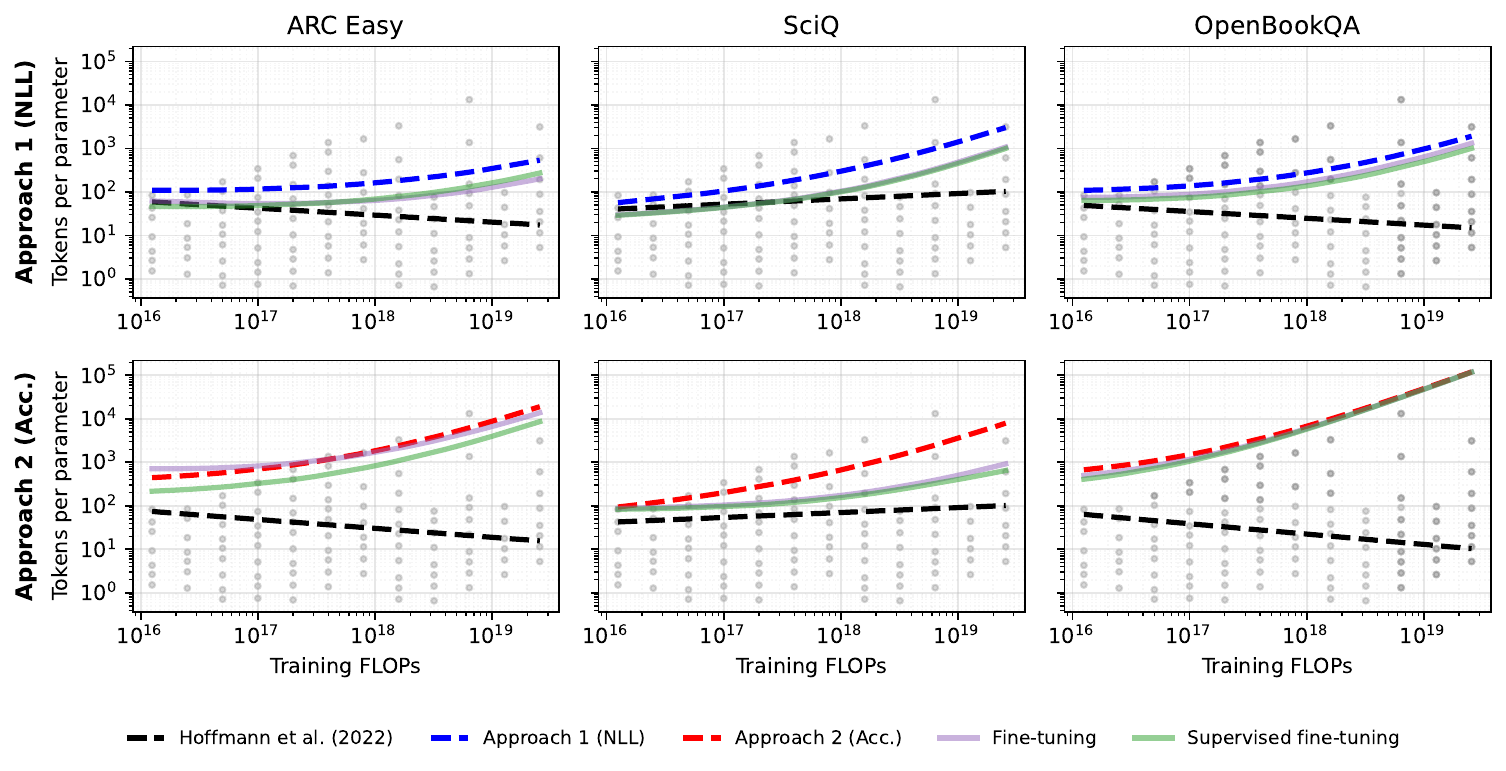}
    \end{tabular}
    \end{center}
    \centering
    \vspace{-8pt}
    \caption{$T^2$ overtraining findings survive post-training. The optimal frontier is slightly subdued compared to base models, which is consistent with~\citet{springer2025overtrained}. 
    }
    \label{fig:rq3}
    \vspace{-2em}
\end{figure}

\textbf{Results.} 
Our results are shown in Figure~\ref{fig:rq3} and Table~\ref{tab:head_to_head}. 
We see in Figure~\ref{fig:rq3} that the optimal frontier continues to shift toward smaller overtrained models with more test-time samples across all three tasks and methods. 
Again, we find that these results are consistent between \textcolor{blue}{Approach~1} and \textcolor{red}{Approach~2}. 
On the other hand, we find that the optimal overtraining recommendation is somewhat subdued compared to $T^2$ on the base models alone, but not enough to shift it back to the original Chinchilla recommendation. 
The finding that it is subdued is consistent with prior work showing that overtrained models are harder to fine-tune~\citep{springer2025overtrained}. 
Finally, we see in Table~\ref{tab:head_to_head} that our best overtrained checkpoints still outperform the Chinchilla optimal checkpoints after post-training, and that performance improves across the board compared to the same analysis on base models in Table~\ref{tab:base_head_to_head}. 
\textbf{This confirms that our findings with $T^2$ scaling persist after post-training.}

\section{Conclusion}
In this work, we have presented $T^2$ scaling laws that jointly optimize model size, training tokens, and the number of repeated samples at test-time under fixed pretraining and inference budgets. 
We find that when test-time compute via repeated sampling is accounted for during pretraining decisions, the optimal model is substantially smaller and more overtrained than what standard Chinchilla scaling prescribes.
This finding is consistent across two complementary modeling approaches: \textcolor{blue}{Approach~1} which models the NLL, and \textcolor{red}{Approach~2} which models the pass@$k$ accuracy directly. 
We validated this across eight real and synthetic downstream tasks, validated that $T^2$ scaling extrapolates to the overtraining regime where its optima are predicted, and that our findings persist after post-training. 
\textbf{Based on our findings, we offer a recommendation to practitioners: if you know your test-time scaling budget with repeated sampling, you should train a smaller model for longer, and $T^2$ scaling offers a blueprint for doing so.} 
In future work, we plan to validate our prescribed overtraining recipes at larger scales, account for transformer-specific inference cost models, and explicitly model the role of post-training in $T^2$ scaling.

\bibliography{refs}
\bibliographystyle{colm2026_conference}

\newpage
\appendix
\section*{Appendix Roadmap}
Our appendix is structured as follows.
We begin with related work in Appendix~\ref{app:related_works}, followed by Appendix~\ref{app:per_task_analysis}, which presents per-task scaling law analyses.
We next turn to experimental details: Appendix~\ref{app:pretraining_details} and Appendix~\ref{app:posttraining_details} describe our pretraining and post-training setups, respectively, while Appendix~\ref{app:evaluation_task} provides descriptions of all evaluation tasks employed in our study.
Finally, Appendix~\ref{app:t^2_scaling} presents the details of our $T^2$ scaling fitting methodology.

\section{Related Work}
\label{app:related_works}
Our work sits at the intersection of three research threads: (i) pretraining scaling laws, (ii) test-time scaling, and (iii) overtrained models.

\subsection{Pretraining Scaling Laws}
\citet{kaplan2020scaling} established that model loss follows predictable power laws as a function of model size and training data.
\citet{hoffmann2022training} (Chinchilla) refined this into \textit{compute-optimal training recipes}, prescribing how model size and token count should scale together under a fixed compute budget.
Recent extensions has broadened the scope of scaling law modeling: studying data quality and quantity~\citep{goyal2024scaling}, incorporating downstream task accuracy~\citep{isik2024scaling, bhagia2024establishing}, decomposing scaling behaviors across knowledge and reasoning skills~\citep{roberts2025compute}, and extending to multimodal settings~\citep{shukor2025scaling}.
These frameworks, however, treat inference as an afterthought---optimizing for a model that is trained once and queried once.
\citet{sardana2024beyond} take a step toward \textit{deployment-aware} scaling by folding inference serving volume into the compute-optimal recipe, yet their analysis is limited to single-pass queries.
We modernize this line of work, where the optimal training decisions must account for both the cost and the compounding performance gains of drawing multiple inference samples.

\subsection{Test-Time Scaling}
Beyond scaling pretraining compute, recent work has increasingly focused on investing computation at \textit{inference time}~\citep{snell2024scaling, zhang2025survey, jaech2024openai, orlanski2025reward}.
This test-time paradigm often focuses on the search for a correct reasoning path rather than the model's inherent knowledge and can broadly be categorized into three regimes: 
(i) \textit{parallel scaling}, which uses consensus through self-consistency~\citep{brown2025large}, or verification over multiple independent responses~\citep{saad2025shrinking};
(ii) \textit{sequential scaling}, which refines reasoning through iterative improvements or hierarchical pruning~\citep{wei2022chain, madaan2023self}; 
and (iii) \textit{internal scaling}, which allows the model to dynamically adjust generation depth based on task difficulty~\citep{jaech2024openai}. 
In this work, we focus on \textit{parallel repeated sampling}---the most common form of test-time scaling---and incorporate pretraining compute budget to jointly optimize allocation decisions.

\subsection{Overtraining}
\citet{hoffmann2022training} (Chinchilla) prescribes a compute-optimal ratio of roughly 20 training tokens per model parameter, yet modern models release routinely deviate from this blueprint by \textit{training smaller models on far more tokens than recommended}.
This deliberate overtraining is motivated by \textit{inference efficiency}: a smaller model costs less per query at deployment.
Recent model families illustrate this trend---Llama-2-7B~\citep{touvron2023llama} was trained on 2T tokens ($\sim$290$\times$ the recommended ratio); 
Google's Gemma-7B~\citep{team2024gemma} was trained on 6T tokens ($\sim$857$\times$), and its successor Gemma~2-9B~\citep{team2024gemma} on 8T tokens ($\sim$889$\times$)---with OLMo~\citep{groeneveld2024olmo} following a similar philosophy.
Our work complements these findings by examining overtraining through a different lens: rather than studying its effect on post-training~\citep{springer2025overtrained}, we show that overtraining is actively \emph{beneficial} when models are deployed with a repeated-sampling inference budget, and we provide a principled framework for determining how much to overtrain given a joint train-and-test compute allocation.
\section{Per-Task Analysis}
\label{app:per_task_analysis}

\begin{figure}[t!]
    \begin{center}
    \begin{tabular}{ccc}
        \includegraphics[width=1.0\linewidth]{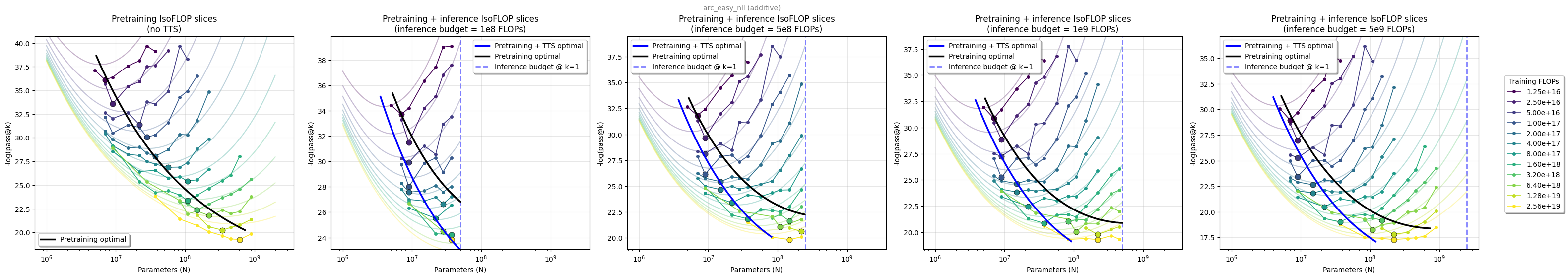} \\
        \includegraphics[width=1.0\linewidth]{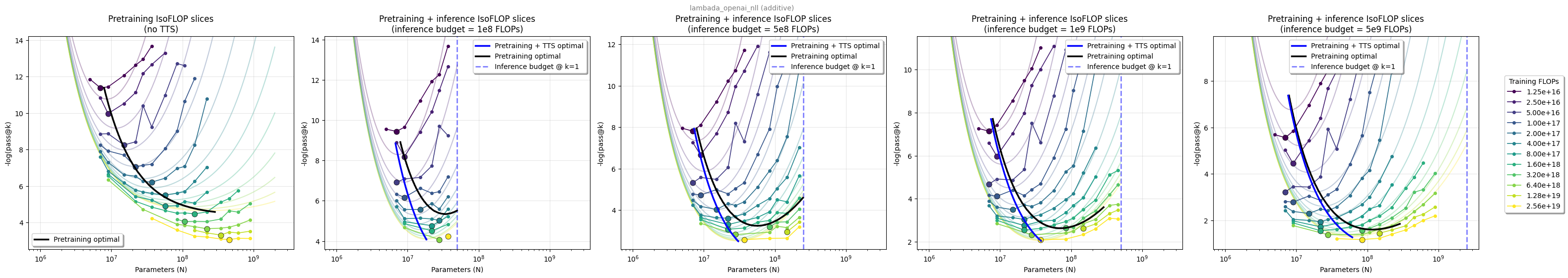} \\
        \includegraphics[width=1.0\linewidth]{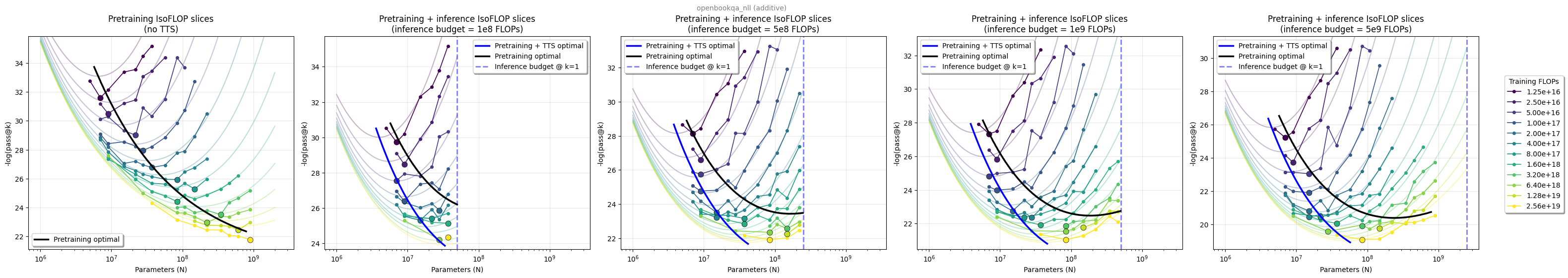} \\
        \includegraphics[width=1.0\linewidth]{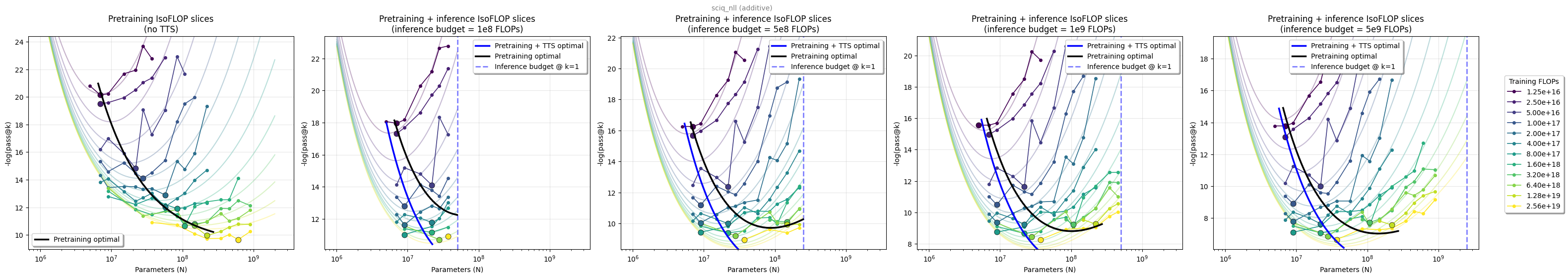} \\
        \includegraphics[width=1.0\linewidth]{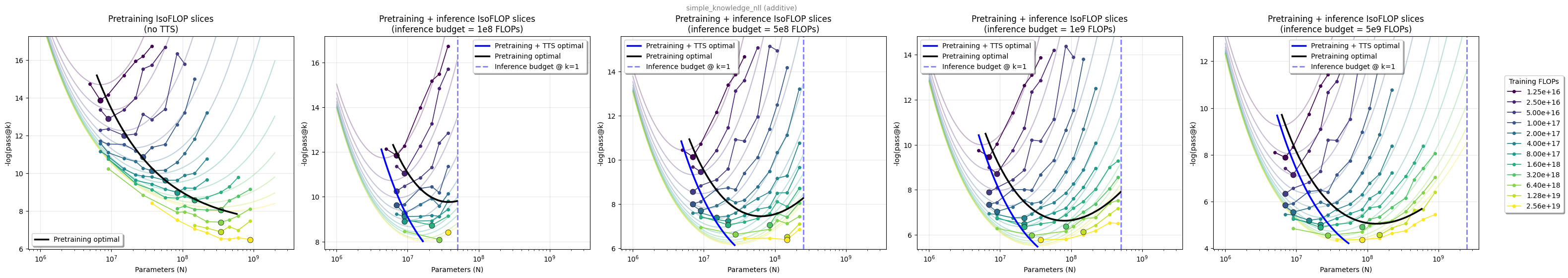} \\
        \includegraphics[width=1.0\linewidth]{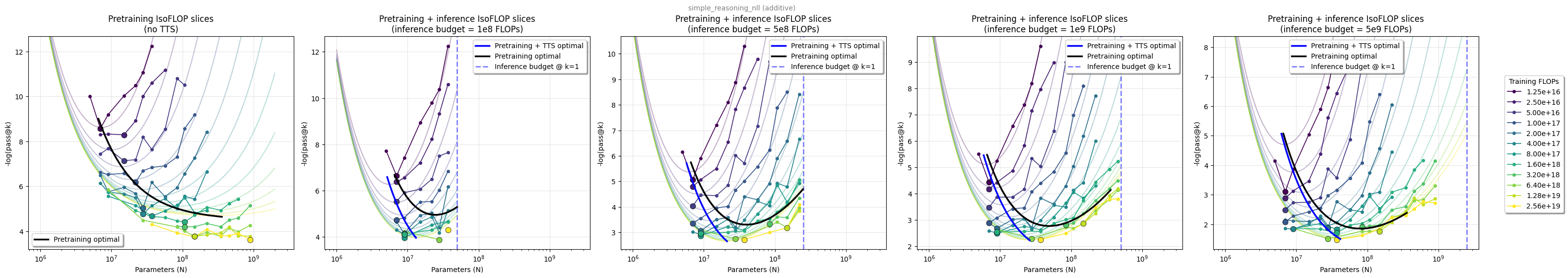} \\
        \includegraphics[width=1.0\linewidth]{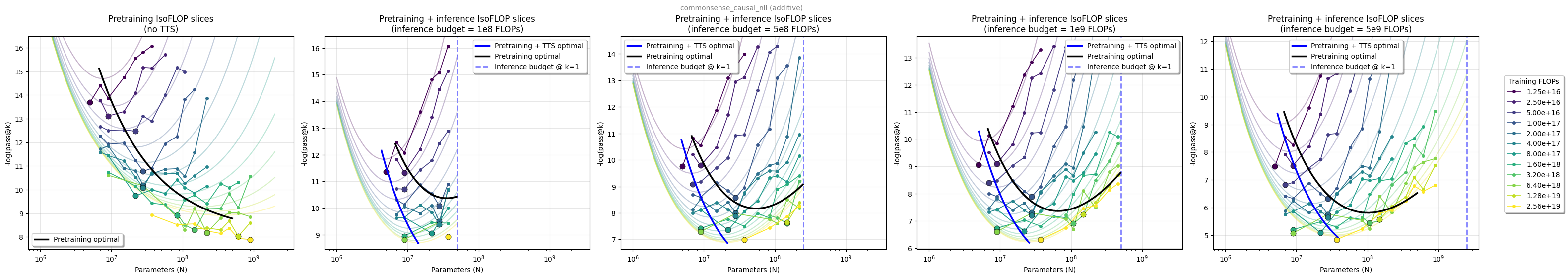} \\
        \includegraphics[width=1.0\linewidth]{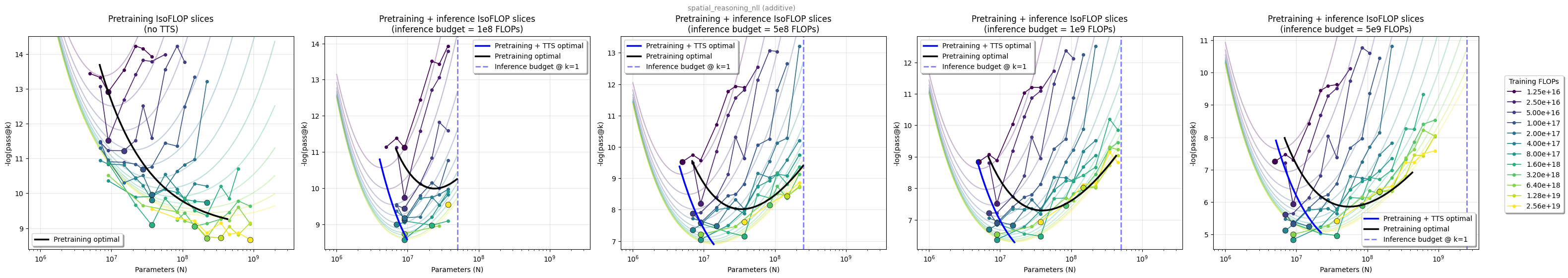} \\
    \end{tabular}
    \end{center}
    \centering
    \caption{
    \textcolor{blue}{Approach~1} IsoFLOP profiles across different scaling budgets for all eight tasks. 
    }
    \label{fig:approach_1_isoflop}
\end{figure}

\begin{figure}[t!]
    \begin{center}
    \begin{tabular}{ccc}
        \includegraphics[width=1.0\linewidth]{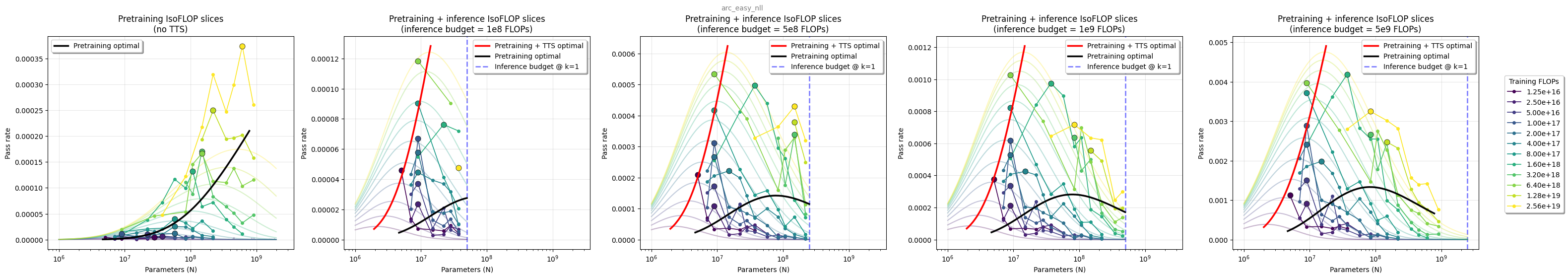} \\
        \includegraphics[width=1.0\linewidth]{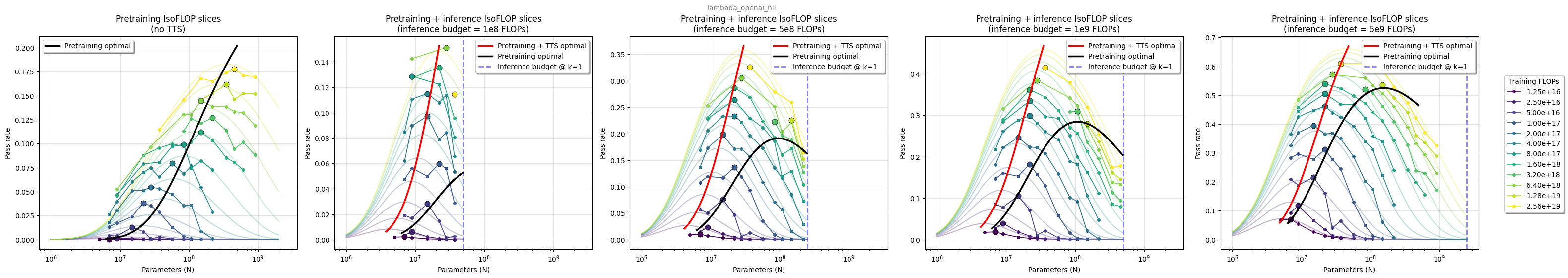} \\
        \includegraphics[width=1.0\linewidth]{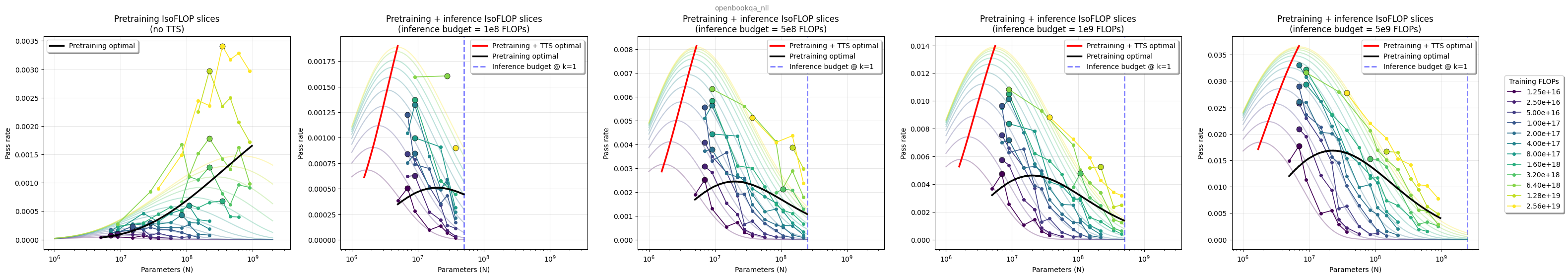} \\
        \includegraphics[width=1.0\linewidth]{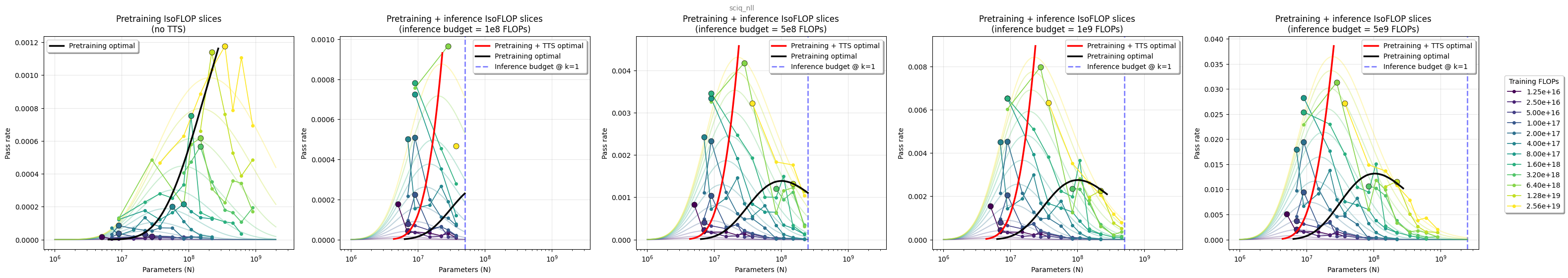} \\
        \includegraphics[width=1.0\linewidth]{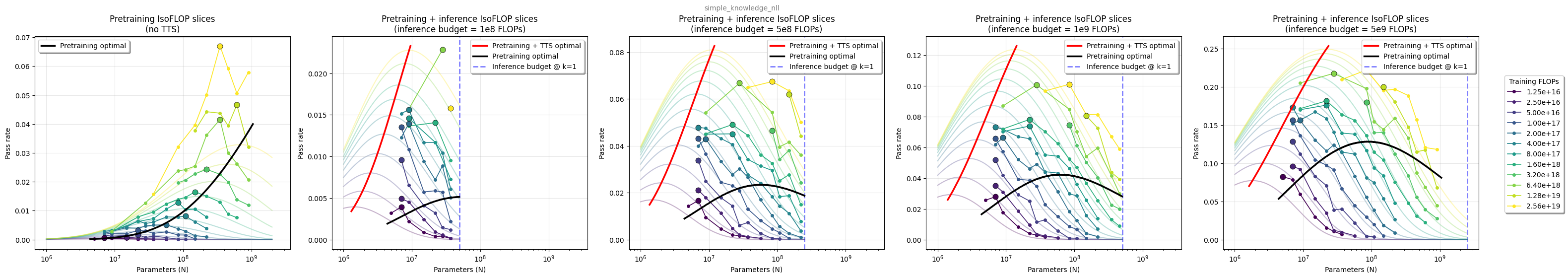} \\
        \includegraphics[width=1.0\linewidth]{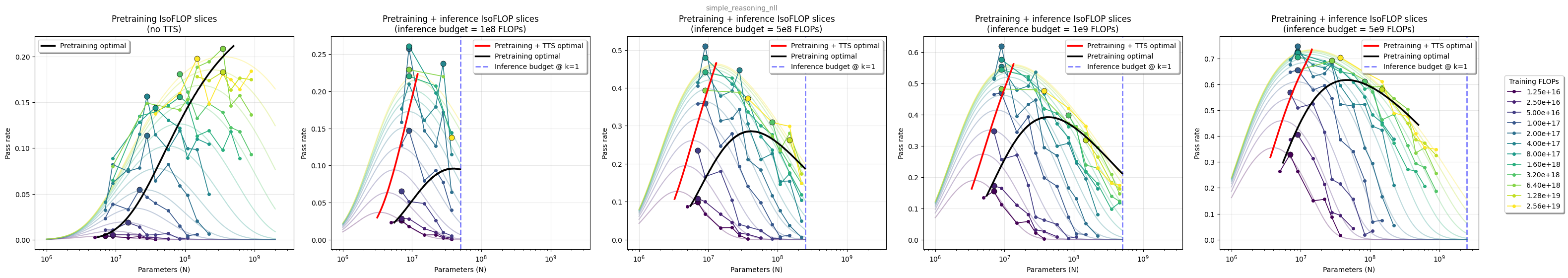} \\
        \includegraphics[width=1.0\linewidth]{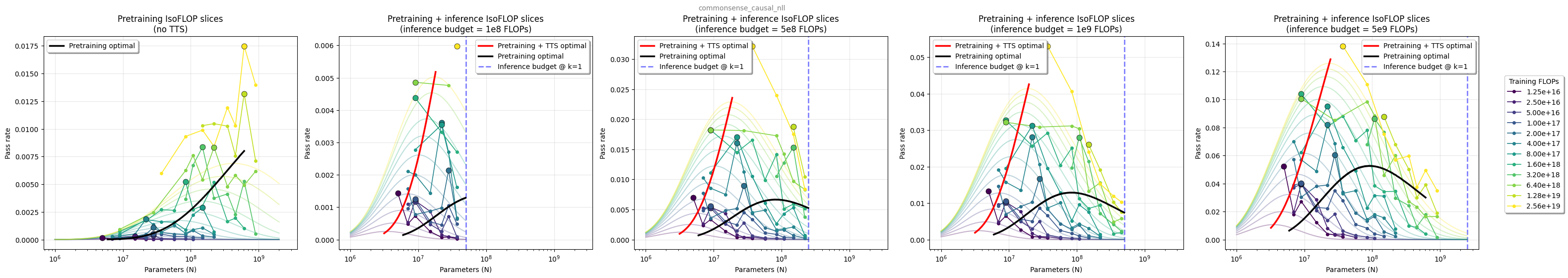} \\
        \includegraphics[width=1.0\linewidth]{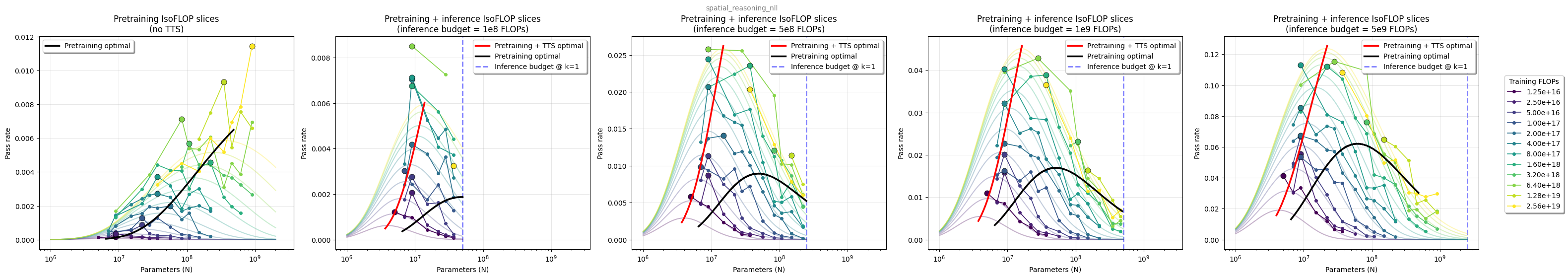} \\
    \end{tabular}
    \end{center}
    \centering
    \caption{
    \textcolor{red}{Approach~2} IsoFLOP profiles across different scaling budgets for all eight tasks. 
    }
    \label{fig:approach_2_isoflop}
\end{figure}

We present isoFLOP profiles for each of the individual tasks in our evaluation suite in Figure~\ref{fig:approach_1_isoflop} for \textcolor{blue}{Approach~1} and Figure~\ref{fig:approach_2_isoflop} for \textcolor{red}{Approach~2} . 
We find that overtraining predictions are relatively stable across inference budgets for both approaches.

\section{Pretraining Details}
\label{app:pretraining_details}

In this section, we provide details of our pretraining setup and scaling grid.

\subsection{Checkpoint Scaling Grid}
\label{app:checkpoint_scaling_grid}
\begin{figure}[t!]
    \begin{center}
    \begin{tabular}{ccc}
        \includegraphics[width=1.0\linewidth]{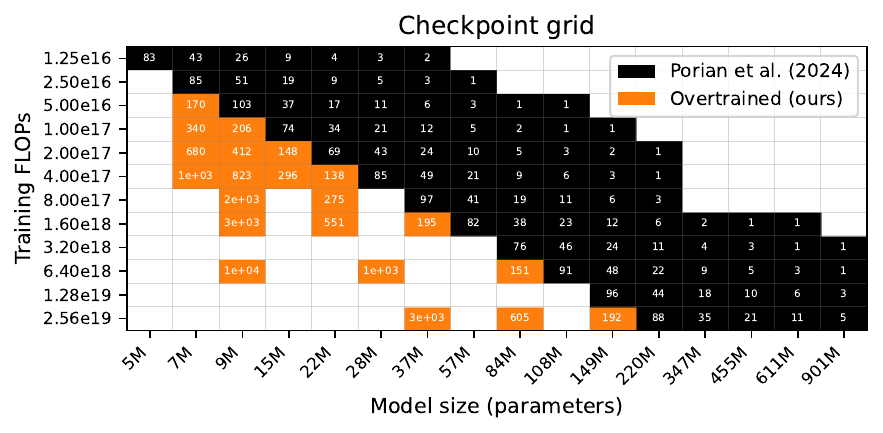}
    \end{tabular}
    \end{center}
    \centering
    \vspace{-8pt}
    \caption{
    Overall checkpoint scaling grid. 
    Each cell reports the number of tokens per parameter.
    Orange cells are overtrained checkpoints we created.
    }
    \label{fig:scaling_grid}
\end{figure}

Figure~\ref{fig:scaling_grid} shows our checkpoint grid, comprising pretrained checkpoints from~\citet{porian2024resolving} alongside additional overtrained checkpoints we pretrained in this work.
Model sizes range from 5M to 901M parameters, and training FLOPs span $1.25 \times 10^{16}$ to $2.56 \times 10^{19}$.
Each cell reports the number of tokens per parameter, which characterizes the degree of overtraining. 
Typically, a suite of Chinchilla scaling checkpoints contains checkpoints at either side of the typical 20 tokens per parameter recommendation derived from \citet{hoffmann2022training}. 
However, since $T^2$ suggests overtraining beyond the available set of checkpoints, we train additional checkpoints at higher tokens per parameter ratios. 
The overtrained checkpoints (shown in orange) are used to validate our forecasts in \S\ref{sec:rq2}.

\subsection{Hyperparameters}
We train our overtrained checkpoints, shown in Figure~\ref{fig:scaling_grid}, from scratch using the OpenLM framework with same fixed hyperparameters used for the Chinchilla-optimal checkpoints from \citet{porian2024resolving}. 
Specifically, we use their \texttt{hparams=base}, \texttt{warmup=short}, \texttt{decay=chinchilla} configuration. 
We use the AdamW optimizer with a learning rate of $3 \times 10^{-3}$, $\beta_1 = 0.9$, $\beta_2 = 0.95$, and a decoupled weight decay of $1 \times 10^{-4}$. 
Training uses a global batch size of 256 sequences of length 2048 tokens, cosine learning rate decay to zero matched to the token budget of each run, and a warmup period equal in tokens to the model's parameter count. 
We apply gradient clipping with a max norm of 1.0, QK-normalization, z-loss with coefficient $10^{-4}$, and train in bfloat16 mixed precision. 
All hyperparameters are held fixed across model sizes, consistent with the base (untuned) configuration of \citet{porian2024resolving}. 
We train on the RefinedWeb dataset with a vocabulary size of 50{,}432. 

\section{Post-Training Details}
\label{app:posttraining_details}

We describe our post-training setup and configurations below.
We employ two variants of post-training: \textbf{(i) standard fine-tuning} and \textbf{(ii) supervised fine-tuning (SFT)}.
Standard fine-tuning follows the conventional next-token prediction objective, computing loss over both the instruction (question) and completion (answer).
SFT, in contrast, computes loss over the completion only, excluding instruction tokens from parameter updates.

We fine-tune on three tasks---ARC Easy~\citep{ai2arc}, SciQ~\citep{SciQ}, and OpenBookQA~\citep{OpenBookQA2018}---covering the full population of pretrained checkpoints, including the overtrained ones.
Each model is trained for 6 epochs until convergence using a batch size of 8 and a constant learning rate of $2 \times 10^{-5}$, after that we evaluate on the respective test set.
All fine-tuning experiments are conducted on 4 NVIDIA A10 GPUs.
Box~\ref{box:data_format} presents the training data format for each task, where the \colorbox{yellow!40}{highlighted} tokens indicate the completion portion used in the SFT loss computation.
Their evaluation follows the same format: we measure negative log-likelihood over the correct answer placed in the \colorbox{yellow!40}{highlighted} placeholder.

\begin{tcolorbox}[
    title=Box 1: Training Data Formats,
    colback=gray!5,
    colframe=gray!50,
    fonttitle=\bfseries,
    label=box:data_format,
    float=t,
]
\small
Each format separates the prompt (plain) from the completion (\colorbox{yellow!40}{highlighted}), which is the only portion used in the SFT loss.
\vspace{4pt}

\textbf{ARC Easy:}
\begin{lstlisting}
Question: {question}\nAnswer:(*@\colorbox{yellow!40}{\texttt{ \{answer\}}}@*)
\end{lstlisting}
\textbf{OpenBookQA:}
\begin{lstlisting}
{question}(*@\colorbox{yellow!40}{\texttt{ \{answer\}}}@*)
\end{lstlisting}
\textbf{SciQ:}
\begin{lstlisting}
{support}\nQuestion: {question}\nAnswer:(*@\colorbox{yellow!40}{\texttt{ \{answer\}}}@*)
\end{lstlisting}
\end{tcolorbox}
\section{Evaluation Tasks}
\label{app:evaluation_task}

Next, we describe the eight downstream tasks used to evaluate $T^2$ scaling, covering both real-world benchmarks and synthetic tasks.
For all tasks, we measure the NLL of each model over the correct answer.

We evaluate on four real-world benchmarks.
\begin{enumerate}[leftmargin=*, itemsep=1pt, topsep=3pt, parsep=0pt]
    \item \textbf{LAMBADA}~\citep{lambada} (OpenAI variant): tests long-range language understanding, where the model must predict the final word of a passage given a broad context.
    \item \textbf{ARC Easy}~\citep{ai2arc}: consists of elementary-level science questions in a four-way multiple choice format, drawn from standardized tests.
    \item \textbf{SciQ}~\citep{SciQ}: contains science exam questions paired with supporting passages, presented in a multiple-choice format.
    \item \textbf{OpenBookQA}~\citep{OpenBookQA2018}: requires multi-step reasoning by combining an open book of core science facts with broader common knowledge, presented as four-way multiple choice questions.
\end{enumerate}

In addition to these four benchmarks, we incorporate four synthetic tasks spanning different domains.
These tasks are designed to evaluate models on (i) simple knowledge recall, (ii) multi-step arithmetic reasoning, (iii) commonsense causal reasoning, and (iv) spatial reasoning.
Each task consists of 1,000 fill-in-the-blank or short-completion questions, generated using GPT-5 and Claude Opus 4.6.
Below, we present representative examples from each task along with their evaluation format.
As in Box~\ref{box:data_format}, the token spans used to compute the NLL are highlighted in each example below.

\begin{tcolorbox}[
    title=Box 2: Commonsense Causal Reasoning,
    colback=cyan!8,
    colframe=cyan!50,
    fonttitle=\bfseries,
    label=box:commonsense_causal_format,
    float=t,
]
\small
\textbf{Example 1:}
\begin{lstlisting}
Grandparents tell stories to grandchildren. Teachers explain
concepts to students. Coaches demonstrate techniques to(*@\colorbox{yellow!40}{\texttt{ players}}@*)
\end{lstlisting}
\textbf{Example 2:}
\begin{lstlisting}
A mother comforts a crying baby. A teacher encourages a
struggling student. A coach motivates a discouraged(*@\colorbox{yellow!40}{\texttt{ player}}@*)
\end{lstlisting}
\end{tcolorbox}

\begin{tcolorbox}[
    title=Box 3: Simple Knowledge Recall,
    colback=cyan!8,
    colframe=cyan!50,
    fonttitle=\bfseries,
    label=box:knowledge_format,
    float=t,
]
\small
\textbf{Example 1:}
\begin{lstlisting}
The capital of Egypt is(*@\colorbox{yellow!40}{\texttt{ Cairo}}@*)
\end{lstlisting}
\textbf{Example 2:}
\begin{lstlisting}
The fifth taste is(*@\colorbox{yellow!40}{\texttt{ umami}}@*)
\end{lstlisting}
\end{tcolorbox}

\begin{tcolorbox}[
    title=Box 4: Multi-Step Arithmetic Reasoning,
    colback=cyan!8,
    colframe=cyan!50,
    fonttitle=\bfseries,
    label=box:reasoning_format,
    float=t,
]
\small
\textbf{Example 1:}
\begin{lstlisting}
I have 5 toys. I give away 2 toys. Step 1: I started with 5
toys. Step 2: I gave away 2 toys. Step 3: 5 minus 2 equals(*@\colorbox{yellow!40}{\texttt{ 3}}@*)
\end{lstlisting}
\textbf{Example 2:}
\begin{lstlisting}
Pattern: 10, 20, 30, ... This adds 10 each time. After 30
comes(*@\colorbox{yellow!40}{\texttt{ 40}}@*)
\end{lstlisting}
\end{tcolorbox}

\begin{tcolorbox}[
    title=Box 5: Spatial Reasoning,
    colback=cyan!8,
    colframe=cyan!50,
    fonttitle=\bfseries,
    label=box:spatial_format,
    float=t,
]
\small
\textbf{Example 1:}
\begin{lstlisting}
The baby is in the crib. The crib is in the nursery. The
nursery is in the house. So the baby is in the(*@\colorbox{yellow!40}{\texttt{ house}}@*)
\end{lstlisting}
\textbf{Example 2:}
\begin{lstlisting}
The glasses are in the case. The case is in the handbag.
So the glasses are in the(*@\colorbox{yellow!40}{\texttt{ handbag}}@*)
\end{lstlisting}
\end{tcolorbox}
\section{Fitting $T^2$ Scaling}
\label{app:t^2_scaling}

In this section, we describe how each of our $T^2$ approaches are fit to empirical checkpoints. 

\paragraph{Fitting \textcolor{blue}{Approach~1}.}
We fit the seven parameters $(\log A, \log B, \log E, \alpha, \beta, \log G, \gamma)$ of the additive model by minimizing the sum of squared errors (SSE) between predicted and empirical NLL values across all checkpoints and sampled values of $k$. 
We use the L-BFGS-B algorithm with 500 random restarts (each with up to 5{,}000 iterations and a tolerance of $10^{-15}$) and we select the run with the lowest objective value.

\paragraph{Fitting \textcolor{red}{Approach~2}.}
We fit the model in two stages. 
First, we fit the standard Chinchilla scaling model $\widehat{L}(N,D) = E + \frac{A}{N^{\alpha}} + \frac{B}{D^{\beta}}$ to the empirical NLL values of all checkpoints. 
We profile over a grid of 40 candidate $E$ values spaced between $0.01 \cdot \min(\text{NLL})$ and $0.95 \cdot \min(\text{NLL})$; for each, we optimize the remaining four parameters $(\log A, \log B, \alpha, \beta)$ via L-BFGS-B with 50+ random restarts, using inverse-variance weighting across isoFLOP groups. 
Second, we fit the Beta regression parameters. 
The per-question success probability is modeled as $p \sim \text{Beta}(a_{N,D}, b_{N,D})$ where $\mu = a_{N,D}/(a_{N,D}+b_{N,D})$ is a scaled logit link and the concentration $\nu = a_{N,D} + b_{N,D}$ is parameterized as a log link function. 
Together, the five parameters $(\theta_0, \theta_1, \theta_2, \theta_3, \theta_4)$ are fit by minimizing SSE between predicted and empirical pass@$k$ accuracy values over a grid of initializations seeded from a sigmoid baseline, again using L-BFGS-B.

\end{document}